\documentclass[pdflatex,sn-basic]{sn-jnl}

\usepackage{graphicx}%
\usepackage{multirow}%
\usepackage{amsmath,amssymb,amsfonts}%
\usepackage{amsthm}%
\usepackage{mathrsfs}%
\usepackage[title]{appendix}%
\usepackage{xcolor}%
\usepackage{textcomp}%
\usepackage{manyfoot}%
\usepackage{booktabs}%
\usepackage{algorithm}%
\usepackage{algorithmicx}%
\usepackage{algpseudocode}%
\usepackage{listings}%

\theoremstyle{thmstyleone}%
\theoremstyle{thmstyletwo}%

\theoremstyle{thmstylethree}%

\begin{document}

\title[EEG-to-Report]{EEG-to-Report: An Annotation and Feature–Text Framework for Training Language Models on Clinical EEG}


\author*[1]{\fnm{Xuan-The} \sur{Tran}}\email{thetx.vck@vimaru.edu.vn}

\author[2]{\fnm{Le Trung Kien} \sur{Nguyen}}

\affil*[1]{\orgdiv{School of Mechanical Engineering}, \orgname{Vietnam Maritime University}, \orgaddress{\city{Haiphong}, \country{Vietnam}}}

\affil[2]{\orgname{HAISmartlink Lab, ANCHI STE}, \orgaddress{\country{Vietnam}}}


\abstract{Clinical electroencephalography (EEG) reporting remains largely manual and time-consuming, and current EEG software ecosystems are not designed to produce the structured EEG–text supervision needed for training modern language models. Most toolboxes focus on visualization, preprocessing, or event marking, but provide limited support for clinician-centred workflows that simultaneously generate high-quality datasets for AI. We introduce \emph{EEG-to-Report}, a browser-based annotation and feature–text framework that links routine EEG review with the construction of AI-ready datasets. The framework integrates multi-format EEG ingestion with automatic channel standardisation and preprocessing, an interactive multi-channel viewer for drag-based time-range and channel selection, and a multimodal annotation layer that combines typed text with voice notes transcribed via speech-to-text. For each annotated segment, a feature extraction engine computes a standardised set of spectral, temporal, entropy, Hjorth, connectivity, and spike-related descriptors, which are stored together with the corresponding clinical descriptions in a portable JSON schema capturing segment timing, channel context, features and notes. This representation yields aligned feature–text pairs designed to supervise multimodal EEG–language models. As a working report generator, the framework includes an in-application auto-report module that combines an ensemble of convolutional networks with a large language model to draft clinical narratives; training a feature-to-text model on the exported corpus is the natural next step, which we outline as future work. Using pilot annotations, we describe how EEG-to-Report is designed to streamline annotation workflows and produce editable draft reports for review by neurologists, providing a reusable foundation for future EEG–text corpora and automated EEG reporting systems.}




\maketitle

\section{Introduction}

Routine clinical electroencephalography (EEG) remains a cornerstone for the functional assessment of the central nervous system and is indispensable in the diagnostic workup of epilepsy, encephalopathy, altered mental status and critical care monitoring.\citep{Beniczky2017SCORE}
However, clinical EEG interpretation is complex, time-consuming, and heavily dependent on specialized expertise that is unevenly distributed across the world. Misinterpretation of EEG can affect large numbers of patients and contribute to both under- and over-diagnosis, motivating the need for computer-assisted tools that improve the quality, consistency and scalability of EEG assessment and reporting in everyday practice.\citep{Beniczky2017SCORE}

A major step toward standardization was the development of SCORE (Standardized Computer-based Organized Reporting of EEG) and its second international version.\citep{Beniczky2013SCORE,Beniczky2017SCORE}
SCORE provides a structured terminology and graphical user interface through which electroencephalographers select predefined descriptors of background activity, sleep, interictal abnormalities, seizures, and neonatal or critical-care patterns, instead of writing free-text reports. This process automatically generates a structured clinical report and simultaneously stores the selected EEG features in a database for quality assurance, education and research.\citep{Beniczky2017SCORE} 
The SCORE-EEG software has been tested on more than 12\,000 recordings across multiple languages and centers, demonstrating the feasibility of standardized, computer-assisted EEG reporting in clinical workflows.\citep{Beniczky2017SCORE} 

Building on this standardized feature–report representation, recent work has leveraged artificial intelligence (AI) to automate parts of EEG interpretation, ranging from convolutional and deep neural networks that decode raw EEG to systems that detect pathological recordings and derive clinical biomarkers.\citep{Schirrmeister2017Decode,Roy2019DLReview,Bajpai2021Pathology,Geraedts2021Biomarkers} Tveit et al.\ introduced SCORE-AI, a convolutional neural network trained on 30\,493 SCORE-annotated EEGs to distinguish normal from abnormal recordings and further classify abnormal EEGs into clinically relevant categories (epileptiform focal/generalized, nonepileptiform focal/diffuse).\citep{Tveit2023SCOREAI} In addition to general pathology classification, deep architectures have been proposed for targeted clinical tasks, such as state-space models for dementia detection.\citep{tran2024eegssm} 
Validated on three independent test sets comprising 9\,945 EEGs, SCORE-AI achieved diagnostic accuracy comparable to expert readers and showed interrater agreement with human consensus within or exceeding the range of human–human variability.\citep{Tveit2023SCOREAI} 
These results indicate that AI models trained on large, structured EEG databases can scale expert-level interpretation to underserved regions and alleviate workload in high-volume centers.\citep{Tveit2023SCOREAI} Furthermore, similar multimodal fusion methods can combine EEG with other behavioral and physiological features to anticipate human performance and states.\citep{Tran2024Multimodal}

More recently, hybrid systems have begun to combine interpretable quantitative EEG features with modern natural language processing (NLP) models for automated report generation. Tung et al.\ proposed a hybrid AI system that extracts standardized descriptors of EEG background and abnormalities, feeds them into machine learning models, and uses a large language model to generate free-text reports aligned with existing clinical narratives.\citep{Tung2024Hybrid}
Their retrospective study on a large clinical dataset demonstrates that such hybrid pipelines can produce clinically plausible background descriptions and impressions, while maintaining compatibility with existing structured reporting practice and scoring systems.\citep{Tung2024Hybrid}
Together, SCORE, SCORE-AI and hybrid EEG–NLP systems illustrate a progression from standardized human-driven reporting to fully automated classification and, more recently, to AI-supported narrative report generation.

Despite this progress, several important gaps remain. First, most existing systems assume access to proprietary, highly curated SCORE-style databases and are tightly integrated with specific commercial EEG platforms, limiting their adoption in smaller centers, research labs, and resource-constrained hospitals that work with heterogeneous EEG file formats and legacy systems.\citep{Beniczky2017SCORE,Tveit2023SCOREAI} Moreover, widely used open-source EEG toolboxes such as EEGLAB, MNE-Python and FieldTrip are designed primarily for signal processing, visualisation and research analysis rather than for clinician-centred generation of aligned feature--text training data.\citep{Delorme2004EEGLAB,Gramfort2013MNE,Oostenveld2011FieldTrip}
Second, current AI models largely treat the signal-to-interpretation process as a monolithic black box: either raw EEG is mapped directly to coarse labels, or fixed sets of human-defined features are mapped to text reports, with limited tooling for clinicians to inspect, curate, and iteratively refine the training data at the level of individual segments.  
Third, while hybrid systems demonstrate that EEG features can drive language models, there is still a lack of open, end-to-end frameworks that (i) help clinicians annotate multi-format EEG recordings; (ii) automatically extract standardized quantitative features from selected segments; and (iii) export aligned feature–text pairs in an \emph{AI-ready} format for training and evaluating EEG-to-report models.

Building such a framework raises three practical challenges that naive extensions of existing tools do not resolve. First, clinical EEG arrives in many vendor formats with inconsistent channel naming, so a viewer bound to a single format or montage cannot serve as a general ingestion layer; robust format-agnostic import and channel standardisation are prerequisites for any downstream corpus. Second, supervision for language models must be aligned at the level of clinically meaningful segments: mapping whole recordings to a single coarse label, as most classifiers do, discards the temporal and spatial context that a report actually describes, so features and text must be bound together per selected segment. Third, capturing rich narrative descriptions without disrupting clinical workflow is difficult: switching between a separate viewer and a text editor is slow, which is why annotation, quantitative feature computation and multimodal (typed or dictated) text entry must be tightly integrated rather than bolted together.

In this work, we present \emph{EEG-to-Report}, an annotation and feature–text framework designed to address these gaps. Our contribution is not a single algorithm but an integrated, browser-based web application that links interactive EEG visualization, multi-modal annotation (text and voice), automated feature extraction, and AI-ready dataset export in one workflow. Concretely, our main contributions are:
\begin{itemize}
  \item We introduce a vendor-agnostic clinical EEG annotation tool that supports more than ten common EEG file formats, provides segment- and channel-level selection, and offers a unified interface for standardized annotations and structured doctor's notes (Section~\ref{subsec:annotation_workflow}).
  \item We define a reusable \emph{feature–text representation} in which each annotated EEG segment is mapped to a set of quantitative time–frequency and connectivity features together with rich clinical text (typed or transcribed speech), exported as JSON for downstream machine learning (Sections~\ref{subsec:feature_extraction} and~\ref{subsec:json_schema}).
  \item We include a working auto-report module that couples an ensemble of convolutional networks with a large language model to draft clinical reports, and we structure the exported feature--text pairs so that local centres can, as future work, train and evaluate their own EEG-to-report models on their data while keeping raw signals and annotations under institutional control (Section~\ref{subsec:model}).
\end{itemize}

By focusing on tooling and data representation rather than a single benchmark model, EEG-to-Report aims to lower the barrier for developing, auditing, and sharing EEG auto-report systems. We envision this framework as a bridge between standardized human annotation traditions such as SCORE and the emerging ecosystem of foundation models and clinical language models for neurology, enabling iterative, human-in-the-loop development of interpretable EEG AI in both academic and clinical environments.

\section{Results}
\label{sec:results}

We demonstrate EEG-to-Report along three axes: (i) its ability to ingest clinical EEG recordings and support clinician-centred annotation, (ii) the structure and content of the resulting EEG feature--text corpus, and (iii) the draft reports produced by the auto-report module. Unless otherwise stated, all analyses were conducted on the publicly available Siena Scalp EEG Database\citep{Detti2020Siena} (Section~\ref{subsec:data_participants}).

\subsection{Data Ingestion and Annotation Workflow}
\label{subsec:results_annotation}

We first assessed data ingestion on the Siena recordings (EDF format). The application parsed channel labels, sampling rates and basic metadata, and standardised channel names to a 10--20-based convention; non-EEG channels (for example ECG or trigger channels), when present, were detected and excluded from the main viewer while remaining accessible for inspection via the channel-selection dialog.

\begin{figure}[t]
\centering
\includegraphics[width=\textwidth]{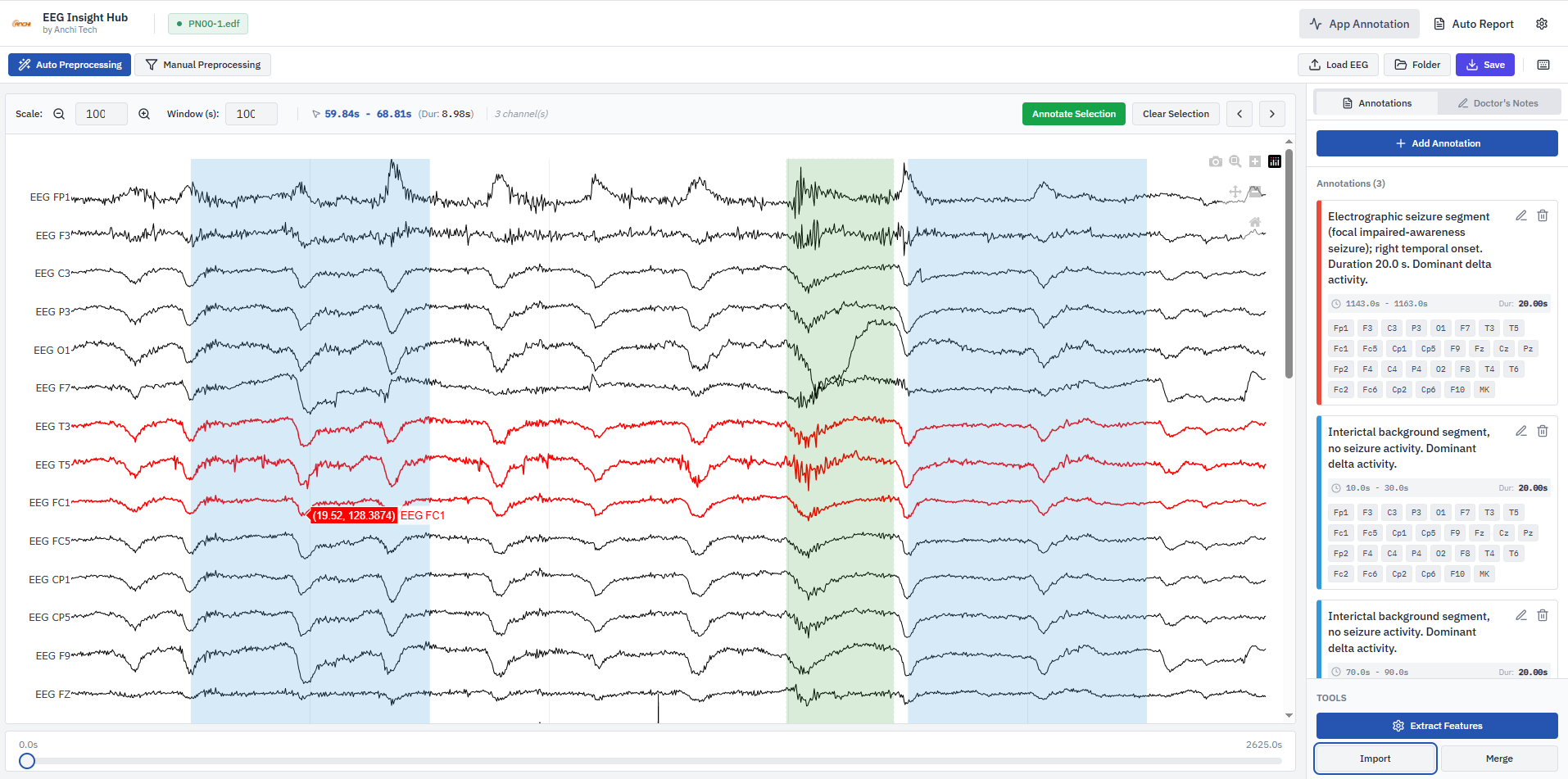}
\caption{The EEG-to-Report annotation interface (Module~1). A drag-selected time range and Ctrl-selected channels are highlighted in the viewer; the annotation dialog captures typed or voice-transcribed descriptions, and saved annotations appear in the timeline for review and navigation.}
\label{fig:ui}
\end{figure}

\begin{figure}[t]
\centering
\includegraphics[width=0.85\textwidth]{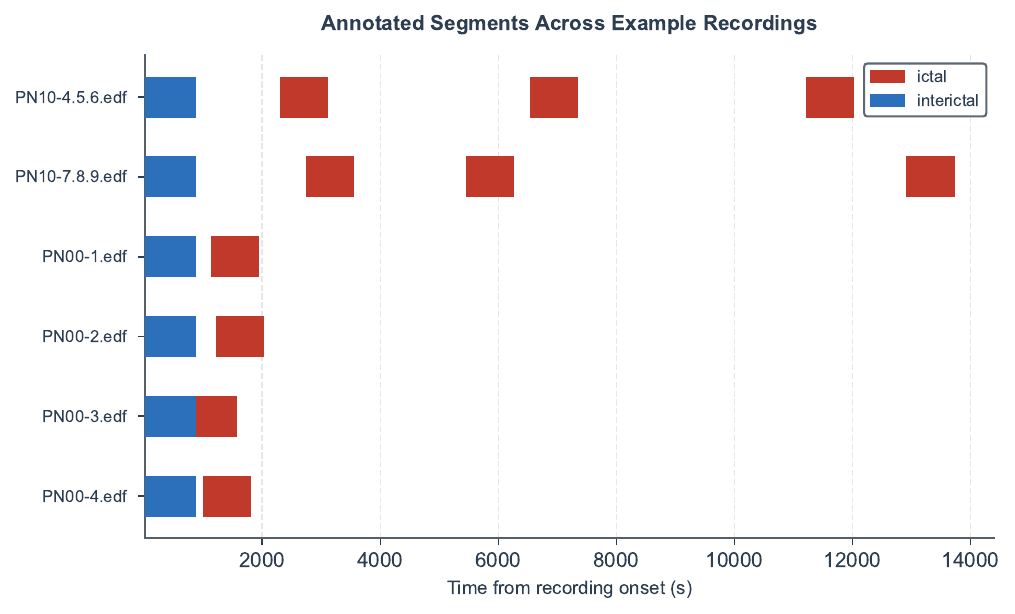}
\caption{Typical annotation timeline across patients in the Siena database, illustrating ictal and interictal segment selection.}
\label{fig:annotation_timeline}
\end{figure}

The interactive viewer (Figure~\ref{fig:ui}) allowed us to navigate full-length recordings and select candidate segments for annotation using the drag-based time-range selection. Guided by the seizure onset/offset markers, we typically worked with time windows of 20.0~s, and also selected shorter segments (e.g. single bursts or discharges) for focal patterns. Across the Siena recordings, the tool was used to create 112 segment-level annotations spanning 36 recordings and 12 patients, covering ictal, peri-ictal and interictal windows. Figure~\ref{fig:annotation_timeline} illustrates typical annotation timelines.

While the system supports speech-to-text input via the Whisper-based voice-annotation interface, in the current demonstration all annotations were entered as direct text (0\% voice). Gathering qualitative clinician feedback on the utility of speech- versus text-based inputs and evaluating speech-to-text accuracy for medical domain terminology are planned for future clinical usability trials.

\subsection{EEG Feature--Text Corpus Statistics}
\label{subsec:results_corpus}

Each saved annotation resulted in a corresponding \emph{segment object} in the JSON dataset, containing segment timing, channel list, the quantitative EEG feature block and the free-text clinical description, with case-level doctor's notes stored separately. Table~\ref{tab:corpus_stats} summarises the corpus statistics.

\begin{table}[t]
\centering
\caption{Summary statistics of the exported feature--text corpus (Siena Scalp EEG Database). Values to be completed from the JSON export.}
\label{tab:corpus_stats}
\begin{tabular}{@{}lr@{}}
\toprule
\textbf{Property} & \textbf{Value} \\
\midrule
Recordings annotated                & 36 \\
Patients                            & 12 \\
Total annotated segments           & 112 \\
Segment duration (s), median [min--max] & 20.0 [20.0--20.0] \\
Channels per segment, median        & 36 \\
Feature-vector length per segment   & 1322 \\
Description length (tokens), median [min--max] & 9 [9--17] \\
Vocabulary size                     & 26 \\
Annotations via voice / text (\%)   & 0 / 100 \\
\bottomrule
\end{tabular}
\end{table}

Segment durations ranged from 20.0 to 20.0~s (median 20.0~s), reflecting both brief paroxysmal events and longer background samples. The feature extraction engine computed, for each segment and channel, band powers in delta, theta, alpha, beta and gamma ranges, basic statistics (mean and standard deviation), Hjorth mobility and complexity, Shannon and approximate entropy, spike counts based on a z-score threshold, and pairwise channel coherence measures. For typical montages with 36 channels, this resulted in feature vectors of length 1322 per segment.

\begin{figure}[t]
\centering
\includegraphics[width=0.8\textwidth]{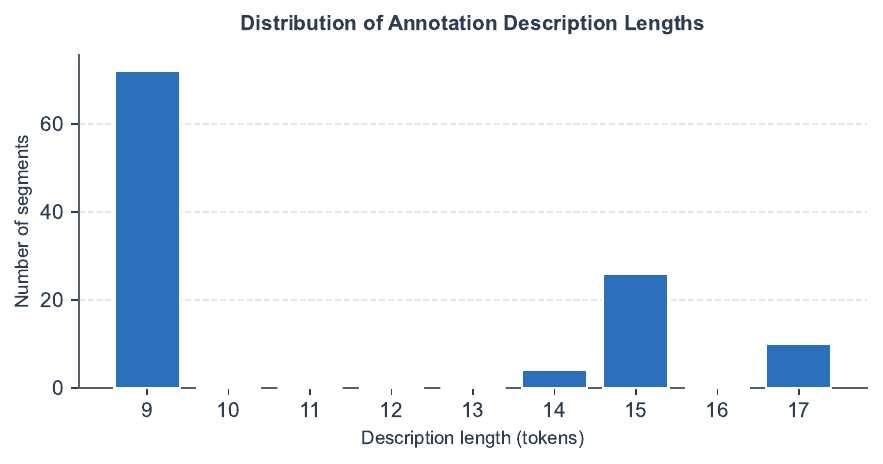}
\caption{Distribution of narrative description lengths (in tokens) across the exported Siena corpus.}
\label{fig:text_len_hist}
\end{figure}

On the text side, each annotation carried a free-text clinical description, complemented at the case level by the doctor's notes. Because the Siena database does not contain narrative clinical descriptions, the text of each segment was seeded programmatically from the dataset's seizure annotations (ictal versus interictal status, and, where available, seizure type and localisation) together with the computed features; these seed descriptions are intended to be reviewed and refined by clinicians within EEG-to-Report rather than treated as expert-authored narratives. To characterise the linguistic content, we computed basic text statistics including token counts, vocabulary size and part-of-speech distributions. Narrative descriptions typically consisted of 9--17 tokens (median 9), with a vocabulary dominated by EEG-specific terminology and common clinical modifiers (for example, ``seizure'', ``interictal'', ``background'', ``onset'', ``focal''). Figure~\ref{fig:text_len_hist} shows the distribution of annotation lengths.

The schema also supports multi-expert annotation: recordings annotated independently by two or more clinicians are stored as separate annotation objects that can later be consolidated. The present single-annotator demonstration does not exercise this capability, but it provides a foundation for future multi-expert studies of inter-rater variability and for training models that account for annotation uncertainty.

\subsection{Auto-Report Generation}
\label{subsec:results_model}

We ran the auto-report module on the referential Siena recordings, where it estimates quantitative descriptors from each recording with the convolutional ensemble and prompts a large language model to generate a narrative (Section~\ref{subsec:model}), and we inspected the resulting drafts.

\begin{figure}[t]
\centering
\includegraphics[width=0.95\textwidth]{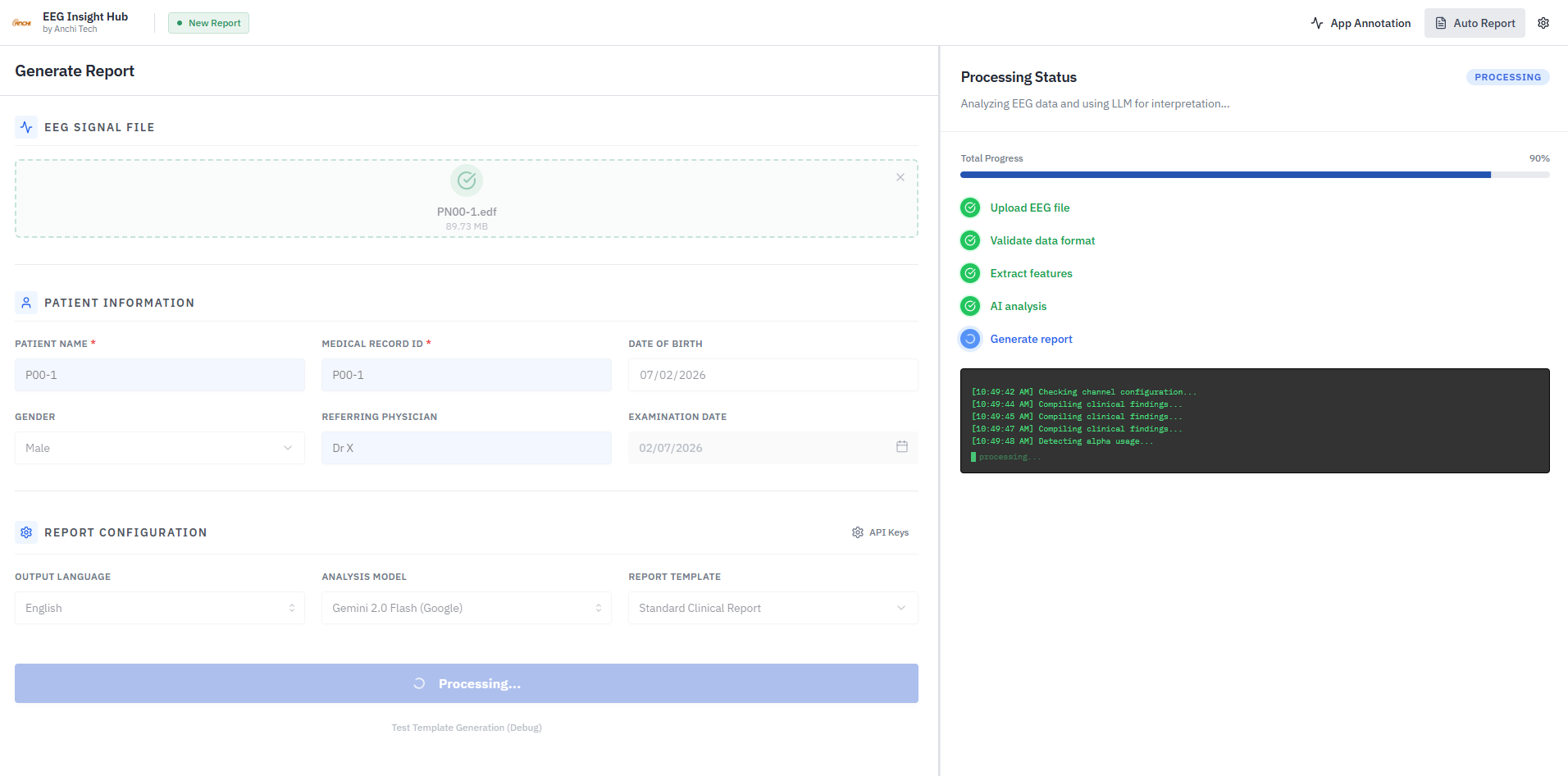}
\caption{The auto-report generation interface in the web application, showing the patient information inputs, the analysis model configuration, and the real-time processing log panel.}
\label{fig:qual_examples}
\end{figure}

\begin{figure}[t]
\centering
\includegraphics[width=0.7\textwidth]{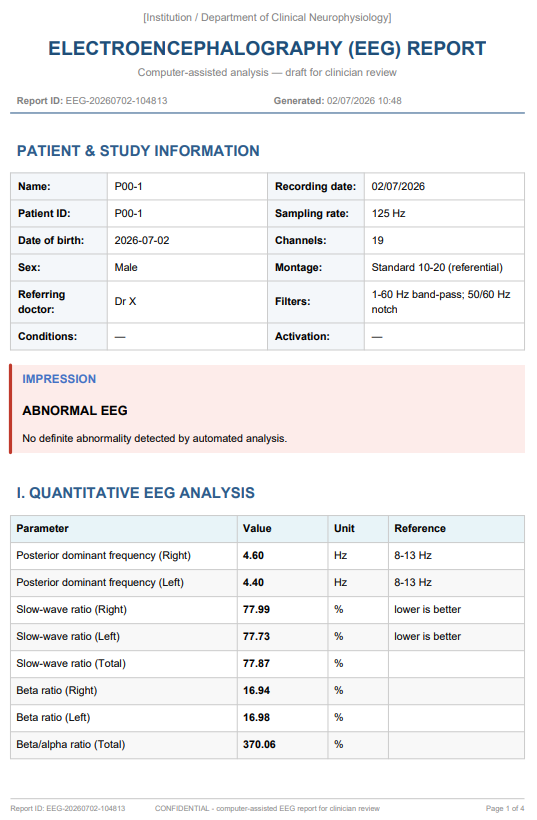}
\caption{The first page of the generated clinical report PDF, showing patient and study metadata, calculated quantitative EEG descriptors (posterior dominant frequencies, slow-wave ratios, beta ratios), and clinical impression. The report generation pipeline and layout are based on the work of Tung et al. \citep{Tung2024Hybrid}.}
\label{fig:generated_report}
\end{figure}

Because the public dataset does not include paired ground-truth clinical reports, we assessed the generated drafts qualitatively rather than with reference-based text-similarity metrics (BLEU, ROUGE-L, BERTScore), which would require a corpus of paired reports and are left to future work. Specifically, 3 expert readers rated the clinical plausibility of the drafts and judged the extent to which they capture core attributes such as background organisation, symmetry, predominant frequency and lateralisation of focal abnormalities. The auto-report interface is shown in Figure~\ref{fig:qual_examples}, and an example of the first page of the generated clinical report PDF is illustrated in Figure~\ref{fig:generated_report}.

Conducting systematic qualitative reviews to categorise failure modes of the generated text (e.g., omission of subtle findings, over-generalised clinical phrasing, or handling of rare institutional terms) and formally assessing clinician usability ratings during real-time review are left to future research.

\subsection{Planned Corpus-Based Analyses}
\label{subsec:results_ablation}

Because the corpus-trained feature-to-text route is future work (Section~\ref{subsec:model}), we outline the sensitivity analyses it will require rather than reporting results here. To characterise the contribution of different feature groups, ablation experiments will train the model with subsets of the full feature vector---removing connectivity features (coherence), removing spectral features, and retaining spectral and basic temporal features alone---to identify which groups contribute most to describing background rhythms, slow-wave abnormalities and finer-grained spatial patterns that depend on coherence and spike counts.

We will also examine the effect of training-set size by subsampling the corpus at different scales, to quantify how text-similarity metrics vary with the number of annotated segments. Such an analysis would indicate the value of tools like EEG-to-Report that can incrementally grow local feature--text corpora over time as part of routine clinical work.

\section{Discussion}
\label{sec:discussion}

In this work, we introduced \emph{EEG-to-Report}, an annotation and feature--text framework that couples multi-format EEG ingestion, clinician-centred segment selection, automated quantitative feature extraction and language-model-based report generation within a single web application. Rather than proposing yet another black-box classifier, our primary contribution is infrastructural: a vendor-agnostic workflow and data representation that make it easier for local centres to build, curate and exploit EEG feature--text corpora with language models, while keeping clinicians in the loop.

\subsection{From Standardised Reporting to AI-Ready Feature--Text Corpora}

The development of SCORE has shown that structured computer-based reporting can harmonise EEG terminology, reduce free-text variability and simultaneously populate databases with machine-readable descriptors selected by experts.\citep{Beniczky2017SCORE}
Subsequent work, such as SCORE-AI, leveraged these structured reports to train large-scale deep learning models that classify EEGs as normal or abnormal and identify major patterns.\citep{Tveit2023SCOREAI}
More recently, hybrid systems have used quantitative EEG features and large language models to generate background descriptions and impressions that mimic human narratives.\citep{Tung2024Hybrid}

EEG-to-Report extends this trajectory in two ways. First, it targets heterogeneous clinical and research environments where SCORE-style infrastructure and proprietary databases may not be available. By supporting more than ten common EEG file formats and relying on open-source libraries for signal processing, our framework allows smaller centres and research groups to accumulate structured annotations across legacy and modern devices without changing their acquisition systems. Second, it moves beyond predefined checklists by aligning quantitative features with rich narrative text at the segment level. Instead of relying solely on predefined categorical descriptors, clinicians can describe patterns in their own words (typed or dictated), while the system automatically binds these descriptions to a comprehensive set of time–frequency and connectivity features. This design preserves the benefits of standardisation (through consistent feature sets and schema) while accommodating the flexibility of free-text narratives that are essential for nuanced clinical communication.

\subsection{A Clinician-Centred, Human-in-the-Loop Workflow}

A central design choice in EEG-to-Report is to keep clinicians at the centre of the annotation and reporting process. The application does not attempt to replace expert readers; rather, it aims to provide a more efficient environment in which segment selection, description and quantitative feature computation are tightly integrated. By design, this integration is intended to reduce friction compared to workflows where clinicians alternate between an EEG viewer and a separate text editor, particularly when speech-based annotation is used for longer impressions.

Importantly, the framework is explicitly human-in-the-loop at two levels. During corpus creation, clinicians author every segment description and note and decide which segments to annotate and which findings to emphasise, thereby shaping the distribution of training data. During report generation, the auto-report module returns its output as an editable draft document (Word or PDF) that the clinician reviews and revises before it is finalised, rather than a fixed verdict. This stands in contrast to end-to-end systems that output final labels or impressions without an obvious path for user correction or feedback, and aligns with recent calls for AI systems that are transparent, corrigible and integrated into clinical workflows rather than operating as external oracles.\citep{nguyen2024fairadxai}

\subsection{Implications for EEG Auto-Reporting and Neuroinformatics}

From a neuroinformatics perspective, the feature--text schema used by EEG-to-Report may be more important than any specific downstream model. By representing each annotated segment as a JSON object that includes timing, channel context, a structured feature vector and clinical text fields, the framework provides a generic interface between EEG time series and language models. This representation is model-agnostic: while a BART-based encoder--decoder is a natural first choice, other architectures such as T5, LLaMA-based encoders, or multimodal transformers could be plugged in with minimal changes to the upstream data pipeline.

This schema also facilitates multi-centre data sharing and benchmarking. Raw EEG files are often difficult to exchange due to size, privacy and vendor constraints, whereas feature--text objects can be more compact and easier to anonymise; large shared resources such as the Temple University Hospital EEG Corpus illustrate both the value and the effort involved in assembling exchangeable EEG datasets.\citep{Obeid2016TUH} Centres that cannot share raw signals might still contribute feature--text pairs and annotation metadata, enabling collaborative training of EEG-to-report models or cross-site evaluations of model generalisability. Furthermore, the inclusion of multi-expert annotations and explicit merge strategies creates opportunities to study inter-rater variability, which is well documented in EEG interpretation,\citep{Interrater2015Cardiac,Interrater2022Enceph} as well as the broader challenges of inter- and intra-subject variability in general EEG analysis,\citep{tran2026inter} and to design models that account for annotation uncertainty, rather than assuming a single ground truth.

For automated reporting specifically, feature-based models have the potential to produce clinically plausible draft descriptions for a non-trivial fraction of segments. This mirrors developments in medical imaging, where deep learning has been widely applied to automatic radiology report generation.\citep{MedReportReview2025,MedReportDL2025} Even when the generated text is not ready for direct use, it can serve as a scaffold that clinicians refine, potentially reducing documentation time and standardising phrasing across readers. Hybrid approaches that combine feature-based descriptions with structured templates, or that condition language models on both features and higher-level labels (for example, SCORE-like codes), may further improve reliability and readability.

\subsection{Limitations}

Several limitations of the present work warrant discussion. First, our empirical evaluation is preliminary and based on a single institution with a limited number of annotators and patient cases. The tool has not yet been evaluated across different hospitals, acquisition systems, or languages, and its usability and performance may vary with local practice patterns. Larger multi-centre studies are required to assess whether the workflow generalises and to quantify the efficiency gains and potential biases introduced by the system.

Second, the current feature set, while extensive, is still hand-crafted. We deliberately focused on interpretable descriptors such as band powers, entropy measures, Hjorth parameters and coherence, which can be linked back to familiar EEG concepts. However, this design may limit the expressiveness of the model compared to approaches that operate directly on raw waveforms or learn representations end-to-end. In addition, connectivity features and spike detection are computed using relatively simple algorithms; more sophisticated methods (for example, source-space connectivity, network measures or advanced spike detection) could yield richer representations at the cost of increased complexity.

Third, report generation is currently split between segment-level annotation (Module~1) and a whole-recording auto-report demonstrator (Module~2), and we have not yet trained a model that learns to generate narratives directly from the segment-level feature--text corpus. Training such a feature-to-text model, and extending it to hierarchical modelling that synthesises study-level reports (combining background, sleep, interictal and ictal findings) from segment-level outputs, are important next steps.

Fourth, like all language-model-based systems, our approach is susceptible to hallucinations, omissions and stylistic biases, and generated text may be vague or over-generalised. Because the ultimate responsibility for interpretation rests with the clinician, careful interface design and training are needed to ensure that users treat generated text as suggestions rather than authoritative conclusions.

\subsection{Future Directions}

The present framework opens several avenues for future research and development. At the methodological level, the feature--text representation could be extended to include raw or lightly processed waveforms as an additional modality, enabling joint training of encoders that learn task-specific features directly from signals, for example through self-supervised representation learning on unlabelled EEG.\citep{Yang2023SSLSleep} Multi-task learning with auxiliary objectives (for example, predicting abnormality labels alongside text) may improve robustness and provide additional control over generated content.

On the tooling side, EEG-to-Report could be integrated with structured reporting standards such as SCORE or ACNS guidelines. For example, the interface could offer optional mapping from free-text descriptions to standardised codes, or present model-suggested descriptors that clinicians accept or adjust. Integration with electronic health records, while beyond the scope of the current work, is another natural step: structured feature--text objects could be linked to patient metadata, diagnoses and outcomes, enabling longitudinal studies of EEG changes and AI-assisted prognostication.

From a deployment perspective, multi-centre validation and continuous learning are key. Because the framework is designed to run locally, centres can incrementally train and update their own models as new annotations accumulate, while keeping sensitive data under institutional control. Federated learning or model-parameter sharing across sites could further accelerate progress without centralising raw EEG data. Finally, the same infrastructure could be adapted to adjacent domains, such as ambulatory EEG, intensive care EEG, or research-grade EEG in cognitive neuroscience, by tailoring annotation templates and feature sets to domain-specific needs.

\section{Methods}
\label{sec:methods}

In this section, we describe the design of the EEG-to-Report framework, its implementation as a browser-based web application, the EEG preprocessing and feature extraction steps, the annotation workflow and feature--text schema, and the report-generation components together with the planned feature-to-text training on the exported corpus.

\subsection{System Overview}
\label{subsec:system_overview}

EEG-to-Report is implemented as a browser-based web application with a client--server architecture, so that it runs on any modern operating system without local installation and keeps signal processing on the server. The frontend is a single-page application built with React and TypeScript (Vite build system), and the interactive EEG viewer uses Plotly.js for hardware-accelerated multi-channel rendering and range selection. The backend is a Python service built with FastAPI, organised as a modular monolith: EEG processing relies on MNE-Python,\citep{Gramfort2013MNE} quantitative features are computed with NumPy/SciPy, and speech-to-text is provided by a locally hosted Whisper model.\citep{Radford2023Whisper} The two sides communicate over a REST API, and all EEG data and derived artefacts remain under institutional control on the server.

\begin{figure}[t]
\centering
\includegraphics[width=\textwidth]{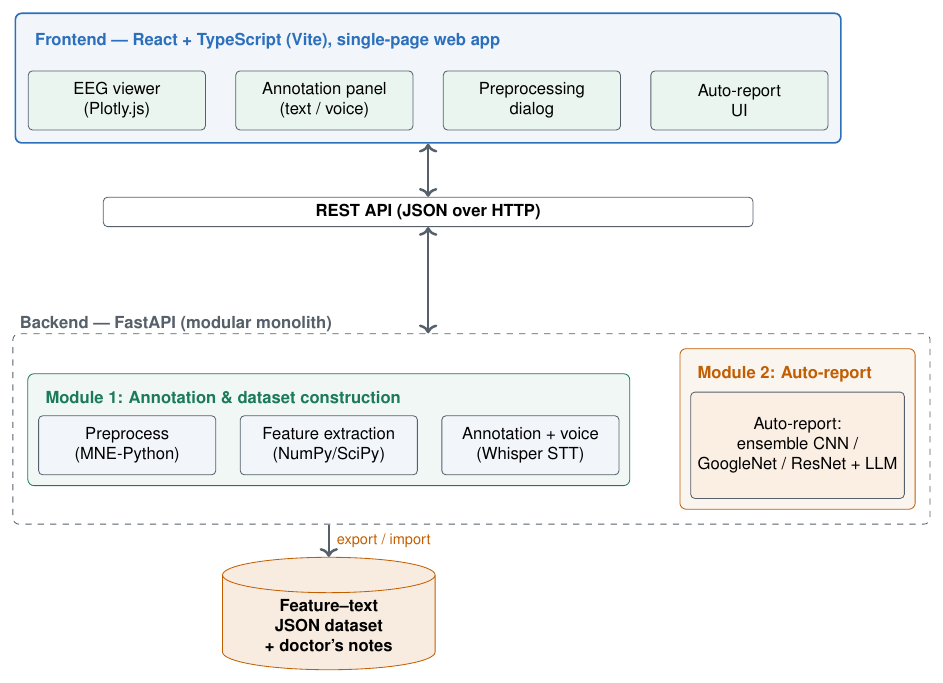}
\caption{System architecture of EEG-to-Report. A React/TypeScript single-page frontend communicates over a REST API with a FastAPI backend organised as a modular monolith. Module~1 (annotation and dataset construction) performs preprocessing, feature extraction and multimodal annotation and exports the aligned feature--text JSON dataset; Module~2 (auto-report) provides an end-to-end report-generation demonstrator. All signal processing runs server-side, keeping raw EEG under institutional control.}
\label{fig:architecture}
\end{figure}

The application is organised into two complementary modules (Figure~\ref{fig:architecture}):
\begin{enumerate}
  \item \textbf{EEG annotation and dataset construction} (the focus of this paper): multi-format EEG import and channel standardisation, an interactive viewer, segment- and channel-level multimodal annotation (typed text and transcribed voice), automatic per-segment feature extraction, and export of aligned feature--text objects in JSON.
  \item \textbf{EEG auto-report}: an end-to-end demonstrator that computes spectral features, applies an ensemble of convolutional networks (CNN, GoogleNet and ResNet) to estimate quantitative descriptors, prompts a large language model to draft a narrative report, and exports it as PDF or Word.
\end{enumerate}

Both modules are decoupled through well-defined REST endpoints, and the annotation module follows a domain-driven layering (presentation, application, domain and infrastructure), so that feature extraction and dataset export can also be scripted independently of the user interface.

\subsection{Data and Participants}
\label{subsec:data_participants}

For the demonstration described in Sections~\ref{sec:results} and \ref{sec:discussion}, we used the publicly available Siena Scalp EEG Database,\citep{Detti2020Siena,Goldberger2000PhysioNet} a de-identified clinical dataset distributed through PhysioNet under the Open Data Commons Attribution License. The database comprises scalp EEG recordings from 14 patients evaluated for epilepsy at the Unit of Neurology and Neurophysiology, University of Siena, acquired with a standard 10--20 referential montage at a sampling rate of 512~Hz and stored in EDF format. Each recording is accompanied by expert-marked seizure onset and offset times, which we used to guide the selection of clinically meaningful segments (for example, ictal, peri-ictal and interictal windows) for annotation; the corresponding seed text descriptions were generated programmatically from these annotations (Section~\ref{subsec:results_corpus}). From this database we used 36 recordings from 12 patients for the demonstration reported below.

Because the dataset is publicly available and fully de-identified at source, no additional ethics approval or informed consent was required for this secondary use; all analyses complied with the PhysioNet data use policy and the Open Data Commons Attribution License of the database.

\subsection{EEG Import and Preprocessing}
\label{subsec:preprocessing}

\subsubsection{Multi-format EEG import}

The application reads the common clinical and research EEG formats through format-specific loaders, including EDF/BDF/GDF (.edf, .bdf, .gdf), EEGLAB (.set), MNE-FIF (.fif), BrainVision (.vhdr), Neuroscan CNT (.cnt) and MATLAB arrays (.mat); additional container formats can be added through the same loader interface. Each file is standardised into an MNE \texttt{Raw} object, preserving channel names, sampling frequency and basic metadata.

During import, channel labels are mapped to a standardised 10--20 naming scheme where possible. Non-EEG channels (for example EOG, EMG, ECG, trigger channels) are detected using name patterns and channel-type hints and are excluded from the default EEG view, but remain available for selection in a channel-management dialog.

\subsubsection{Preprocessing pipeline}

To provide a consistent basis for annotation and feature extraction, EEG-to-Report applies a default preprocessing pipeline that can be customised by the user:
\begin{itemize}
  \item \textbf{Bandpass filtering}: recordings are bandpass-filtered using a zero-phase finite impulse response (FIR) filter with default passband 1--60~Hz. Low- and high-cutoff frequencies can be changed via a preprocessing options dialog.
  \item \textbf{Notch filtering}: line noise at 50 or 60~Hz (and harmonics, if desired) can be removed using an IIR notch filter.
  \item \textbf{Re-referencing}: signals can be re-referenced to the average of all EEG channels or to a user-selected reference channel.
  \item \textbf{Resampling}: data are resampled to a standard frequency (default 125~Hz) for computational efficiency and to harmonise datasets from different acquisition systems.
  \item \textbf{Independent component analysis (ICA)}: an ICA decomposition (typically 15--30 components) can be computed on a copy of the data for artefact removal. Components corresponding to eye movements or other stereotyped artefacts can be identified automatically based on correlation with EOG channels (if present) or manually inspected and marked for removal.
  \item \textbf{AutoReject-based epoch cleaning}: for users who choose to work with fixed-length epochs, an automatic artefact rejection step can be applied using threshold-based methods to mark and remove epochs with excessively large amplitudes.
\end{itemize}

Preprocessing operations are applied to the in-memory \texttt{Raw} object held by the server for the active recording, and the chosen parameters (filter band, notch frequency, resampling rate, reference and ICA settings) are exposed in the interface so that the same pipeline can be reproduced on other recordings.

\subsection{Interactive Visualisation and Annotation Workflow}
\label{subsec:annotation_workflow}

\begin{figure}[t]
\centering
\includegraphics[width=\textwidth]{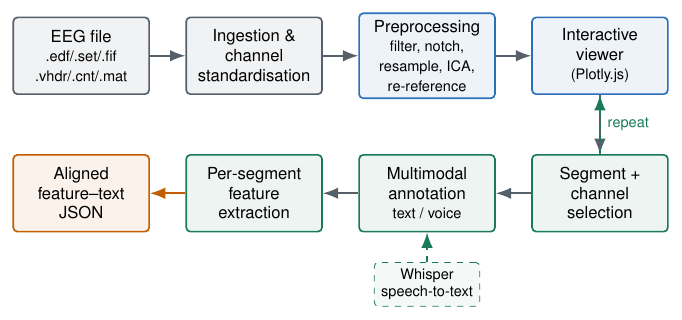}
\caption{Dataset-creation pipeline. A multi-format EEG recording is ingested and channel-standardised, preprocessed, and reviewed in the interactive viewer. The clinician selects a time segment and channels, annotates them with typed text or voice (transcribed by Whisper), and the system computes per-segment quantitative features; each annotated segment is stored as an aligned feature--text pair in the exported JSON.}
\label{fig:pipeline}
\end{figure}

The end-to-end dataset-creation workflow of Module~1---from multi-format ingestion through preprocessing, interactive review and multimodal annotation to per-segment feature extraction and export of the aligned feature--text JSON---is summarised in Figure~\ref{fig:pipeline}.

\subsubsection{EEG viewer}

The Plotly.js-based viewer displays multiple EEG channels as stacked traces over time. Users can adjust the visible time window (default 10~s, range 1--60~s) and vertical scaling, and channels are colour-coded to indicate their selection status. Channel labels are displayed on the left, aligned with their corresponding traces.

Navigation controls include \texttt{Previous} and \texttt{Next} buttons for stepping through the recording, a time-window control, and a timeline overview for jumping across the full duration. The viewer adapts to the number of channels in the current montage.

\subsubsection{Segment and channel selection}

Annotations are created by selecting both a temporal interval and one or more channels:
\begin{enumerate}
  \item The user drags a selection box across the viewer to define a highlighted time range, which is reflected back in real time.
  \item Individual channels are then added to the selection by clicking on their traces, with \texttt{Ctrl+click} for multi-channel selection; if no channel is chosen explicitly, the annotation applies to all displayed channels.
  \item A keyboard shortcut (or the annotation button) opens the annotation dialog, pre-populated with the selected time range and channel list.
\end{enumerate}

Existing annotations are displayed as coloured bars above the time axis, and clicking on a bar moves the viewer to the corresponding time range. An annotation list panel shows all annotations for the current recording, including start and end times, channels, and short text snippets.

\subsubsection{Text and voice annotations}

The annotation dialog provides two tabs:
\begin{itemize}
  \item \textbf{Text annotation}: a multi-line text field for entering narrative descriptions of the selected segment (for example, background organisation, focal slowing, epileptiform discharges).
  \item \textbf{Voice annotation}: a simple audio recorder that captures a short spoken description and streams it to a Whisper-based speech-to-text backend. The transcribed text appears in an editable field, allowing the clinician to correct errors before saving. The original audio file can optionally be stored for later review.
\end{itemize}

Each annotation is stored with a unique identifier and metadata including segment start and end times (in seconds from recording onset), the list of channels, annotation type (text/voice), a display colour and creation/update timestamps.

\subsection{Quantitative Feature Extraction}
\label{subsec:feature_extraction}

For every saved annotation, EEG-to-Report extracts a vector of quantitative features from the selected channels and time interval. Features are computed on preprocessed data using NumPy/SciPy routines and MNE helper functions.

\subsubsection{Spectral features}

Power spectral density (PSD) is estimated using Welch's method with a Hamming window and 50\% overlap. From the PSD, we compute absolute and/or relative band power in the canonical frequency bands for each channel: delta (0.5--4~Hz), theta (4--8~Hz), alpha (8--12~Hz), beta (12--30~Hz), and gamma (30--50~Hz). We also derive the dominant frequency band and, where applicable, peak frequency within the alpha band.

\subsubsection{Time-domain features}

For each channel, we compute: (i) mean amplitude and standard deviation, (ii) Hjorth mobility and complexity,\citep{Hjorth1970} and (iii) linear trend (for optional detrending analysis). These features summarise basic amplitude and variability characteristics of the segment.

\subsubsection{Entropy-based features}

To capture signal irregularity, which has been shown to discriminate clinically relevant EEG states,\citep{Sabeti2009Entropy} we compute the Shannon entropy of the amplitude distribution and the Approximate entropy with embedding dimension $m=2$ and tolerance parameter $r=0.2\sigma$ (where $\sigma$ is the standard deviation of the segment).\citep{Pincus1991ApEn} These measures quantify complexity and predictability of the EEG traces within the selected interval.

\subsubsection{Connectivity and spike features}

Connectivity between channels is characterised using magnitude-squared coherence, computed for all channel pairs and summarised in a symmetric coherence matrix. For storage efficiency, we flatten the upper triangle of the coherence matrix into a feature vector or, optionally, compute summary statistics (for example, average intra-hemispheric and inter-hemispheric coherence).

Spike-related activity is quantified using a simple z-score-based detector: samples exceeding a configurable threshold (default 3 standard deviations) are counted as spikes, and spike counts per channel and per segment are stored as features.

\subsection{Feature--Text Schema and JSON Dataset}
\label{subsec:json_schema}

\begin{figure}[t]
\centering
\includegraphics[width=\textwidth]{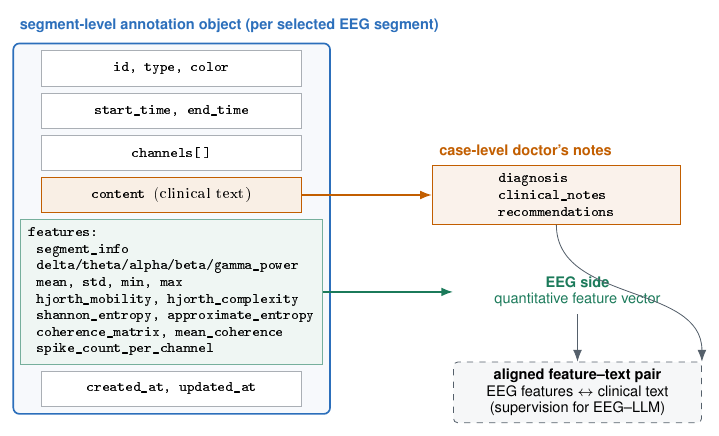}
\caption{Feature--text JSON schema. Each clinician-selected segment is stored as an annotation object containing timing, channels, the free-text clinical description (\texttt{content}) and a structured block of quantitative features (band powers, temporal statistics and Hjorth parameters, entropy measures, coherence and spike counts); case-level doctor's notes are stored separately. The feature block (EEG side) and the clinical text (text side) form an aligned feature--text pair that can supervise multimodal EEG--language models.}
\label{fig:schema}
\end{figure}

All information pertaining to annotated segments is stored in a JSON-based dataset format designed for portability and model training (Figure~\ref{fig:schema}). The exported dataset comprises two parts: a list of \textbf{segment-level annotation objects} and a separate \textbf{case-level doctor's-notes object}.

Each segment-level object includes: (i) a unique segment identifier, (ii) start and end times (seconds from recording onset), (iii) the list of channels included in the segment, (iv) the annotation type (text or transcribed voice) and a display colour, (v) the free-text clinical description (the \texttt{content} field), (vi) the quantitative feature block, structured into named groups (segment info, spectral band powers, temporal statistics and Hjorth parameters, entropy measures, channel coherence, and spike counts), and (vii) creation and update timestamps. The case-level object stores the clinician's overall \texttt{diagnosis}, \texttt{clinical\_notes} and \texttt{recommendations}. Together, the segment description (and, at the case level, the doctor's notes) provides the \emph{text} side and the feature block provides the \emph{EEG} side of each aligned feature--text pair.

EEG-to-Report can import and export these JSON files, and can consolidate temporally adjacent annotations using a configurable time-gap threshold (concatenating their descriptions and taking the union of their channels). This provides a foundation for combining annotations; richer multi-expert conflict-resolution strategies are a planned extension (Section~\ref{sec:discussion}).

\subsection{Report Generation and Feature-to-Text Modelling}
\label{subsec:model}

\begin{figure}[t]
\centering
\includegraphics[width=\textwidth]{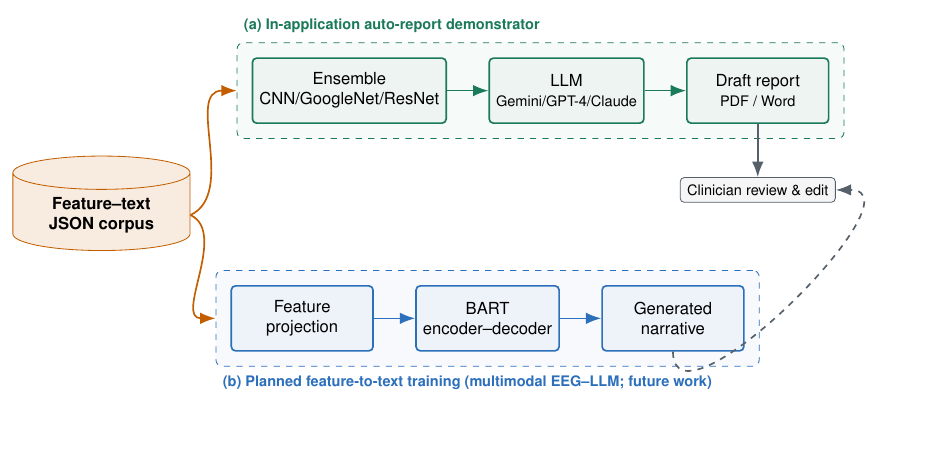}
\caption{From the feature--text corpus to language models. (a) The in-application auto-report demonstrator (available today) estimates quantitative descriptors with an ensemble of convolutional networks and prompts a large language model to draft a report exported as PDF or Word. (b) A planned feature-to-text route (future work) would project the segment feature vector and fine-tune a BART encoder--decoder to generate narratives, showing how the exported corpus is designed to supervise multimodal EEG--language models. In both paths the draft is reviewed and edited by a clinician.}
\label{fig:llm}
\end{figure}

The framework connects EEG features to language models in two ways: an in-application auto-report demonstrator that is available today, and a planned feature-to-text training route that will consume the exported dataset to build multimodal EEG--language models (Figure~\ref{fig:llm}).

\subsubsection{In-application auto-report demonstrator}

The auto-report module illustrates end-to-end report generation inside the application. Power spectral density features from posterior channels are normalised and passed to an ensemble of three convolutional architectures---a plain CNN, an Inception-style network (GoogleNet) and a residual network (ResNet)\citep{Schirrmeister2017Decode}---whose predictions are averaged to estimate quantitative descriptors of the recording. These descriptors are inserted into a structured prompt and submitted to a large language model, with support for Google Gemini, OpenAI GPT-4 and Anthropic Claude and automatic fallback between providers, which drafts a clinical-style report; the draft is presented for clinician review and exported as PDF or Word. This module is based on the report generation pipeline and code structure proposed by Tung et al. \citep{Tung2024Hybrid}, serving as a working reference for the kind of downstream system the exported corpus is intended to support.

\subsubsection{Planned feature-to-text modelling on the exported corpus}

The exported corpus is designed to be directly usable for training multimodal EEG--language models offline, independently of the deployed application; here we outline the intended setup, which we leave as future work. For each segment, the quantitative feature vector would be normalised (for example, z-scored per feature across the training set) and, where necessary, reduced in dimensionality; the resulting vector of length $D$ would be partitioned into $L$ groups (for example, by feature type), yielding an input sequence with $d$-dimensional embeddings ($D \approx L \times d$), and a learned linear projection would map each group into the encoder hidden space.

A natural choice for this task is a BART-based encoder--decoder, following the transformer sequence-to-sequence paradigm.\citep{Vaswani2017Attention,Lewis2020BART} The encoder would consume the projected feature sequence, the decoder would generate the corresponding narrative description token by token, and parameters would be initialised from a pre-trained BART checkpoint and fine-tuned on the feature--text pairs exported by EEG-to-Report. Training would use teacher forcing with a cross-entropy loss, the dataset would be split at the patient level to prevent information leakage, and text would be tokenised with the BART tokenizer. A full implementation, hyperparameter search and evaluation of this route are beyond the scope of the present paper, which focuses on the tool and the dataset representation that make such training possible.

\subsection{Evaluation Protocol}
\label{subsec:evaluation}

We assessed EEG-to-Report along two dimensions:

\paragraph{Data ingestion and annotation.}
We verified that the application could import the EEG recordings from the Siena Scalp EEG Database and correctly standardise channel labels and metadata. To characterise annotation behaviour, we computed statistics on the number of annotations per recording, segment duration, and the distribution of text lengths and vocabulary. We also documented the proportion of annotations created via voice versus text input.

\paragraph{Report generation.}
For the auto-report module, formal qualitative clinical assessments by expert raters (scoring drafts on a scale of acceptability and identifying error categories) are planned as part of future work. Because the dataset contains no paired reference reports, reference-based text-similarity metrics (such as BLEU, ROUGE-L, or BERTScore), alongside ablation and subsampling analyses of a corpus-trained feature-to-text model, are similarly described as planned future work (Section~\ref{subsec:results_ablation}).

This protocol is designed to demonstrate the technical feasibility of the EEG-to-Report pipeline on public clinical data, while acknowledging that dedicated usability studies and larger multi-centre evaluations will be required to fully characterise workflow benefits, performance and generalisability.

\backmatter

\bmhead{Acknowledgements}
The authors thank the contributors of the Siena Scalp EEG Database and the PhysioNet platform for making the data used in this study publicly available.

\section*{Declarations}

\begin{itemize}
  \item \textbf{Competing interests.} The authors declare that they have no competing interests.
  \item \textbf{Ethics approval.} This study used the publicly available, fully de-identified Siena Scalp EEG Database\citep{Detti2020Siena,Goldberger2000PhysioNet} and did not involve any new collection of data from human participants. Ethics approval and informed consent were therefore not required for this secondary analysis; all use complied with the database licence and the PhysioNet data use policy.
  \item \textbf{Consent to participate.} Not applicable (secondary use of publicly available, de-identified data).
  \item \textbf{Consent for publication.} Not applicable.
  \item \textbf{Data availability.} The EEG recordings analysed in this study are publicly available from the Siena Scalp EEG Database on PhysioNet (\url{https://doi.org/10.13026/5d4a-j060}).
  \item \textbf{Code availability.} The source code of the EEG-to-Report application is available from the corresponding author upon reasonable request.
  \item \textbf{Author contributions.} X.-T.T. conceptualised the study, developed the core software framework, curated the data, and wrote the manuscript. L.T.K.N. optimised the user interface, implemented backend functions, and performed dataset preprocessing. All authors reviewed and approved the final manuscript.
\end{itemize}

\bibliography{sn-bibliography}

\end{document}